\documentclass[10pt,a4paper]{article}
\usepackage{hyplm_report}
\renewcommand{\today}{2026-10-12}
\hypersetup{pdftitle={HyPLM Technical Report},
 pdfauthor={Hyra Team},
 pdfsubject={HyPLM-Flash, an efficient and open protein language model},
 pdfkeywords={HyPLM, protein language model, masked language modeling, efficient inference}}
\title{HyPLM Technical Report}
\author{Hyra Team}
\begin{document}
\maketitle

\begin{abstract}
Protein language models (PLMs) are standard tools for variant-effect prediction, protein annotation and
design, but the compute needed to train and serve them keeps growing.
We present HyPLM-Flash, an efficient and open PLM trained by Hyra 2.0, the research agent developed by
our team.
HyPLM-Flash is a 650-million-parameter masked language model trained on 726 billion tokens, 0.73 times
the data of ESM-2 650M and a fifth to an eighth of the data of E1-600M and ESM~C 600M.
It runs on an inference engine with fused kernels and INT8 matrix multiplication, embedding about 130,000
residues per second on a single NVIDIA H20 GPU, 1.9 to 2.7 times faster than other 600M-class encoders.
In a matched evaluation against 28 public checkpoints on 32 tasks, HyPLM-Flash reaches a ProteinGym
zero-shot Spearman correlation of 0.409, statistically tied with ESM-2 650M, ESM~C 600M and E1-600M, and
the highest organismal-fitness score among self-supervised single-sequence models.
It gives the best long-range contact precision of all models evaluated, including ESM-2 15B and ESM~C 6B,
and a lower masked-residue loss than ESM-2 650M, ESM~C 600M and E1-600M on held-out sequences and on
recently released PDB chains.
Matching state-of-the-art accuracy at a fraction of the training and inference cost, HyPLM-Flash sets
a new Pareto frontier for protein language models.
We have released the model weights and the inference engine at
\url{https://github.com/Tencent-Hunyuan/Hyra-results/tree/main/AI4Science/hyplm_flash}.

\end{abstract}

\begin{figure}[H]
\centering\includegraphics[width=\linewidth]{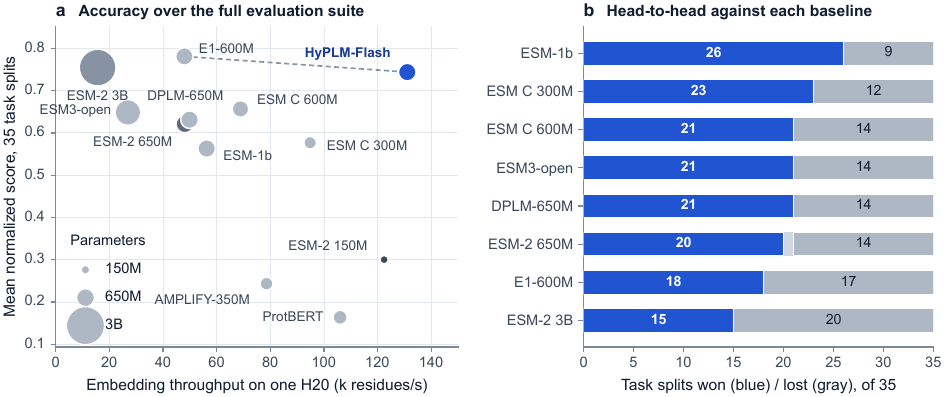}
\caption{\textbf{Accuracy and inference speed of protein language models.}
\textbf{a,} Mean score over the 35 task splits of our evaluation suite on which all plotted models
have a value, against embedding throughput on one NVIDIA H20 GPU. Scores are min--max normalized on each
split across the twelve models. Throughput is measured with the INT8 engine for HyPLM-Flash and with the
reference implementations for the baselines. The dashed line connects Pareto-optimal models; bubble area is
proportional to parameter count.
\textbf{b,} Number of these splits that HyPLM-Flash wins (blue) and loses (gray) against each baseline.}
\label{fig:teaser}
\end{figure}

\section{Introduction}\label{sec:intro}
Protein language models (PLMs) learn the statistics of natural sequences by reconstructing masked residues,
and the representations they learn capture structure, function and fitness without
supervision~\citep{rives2021biological,rao2020transformer,lin2023evolutionary}.
Encoders with a few hundred million parameters have become everyday tools for variant-effect prediction,
protein annotation and fine-tuning, and they are increasingly run over entire proteomes and metagenomes.
Their cost therefore matters.
Recent encoders of the 600M class were trained on 4 to 6 trillion tokens~\citep{esmc2024,jain2025e1}, four
to six times the data of ESM-2, and on an NVIDIA H20 their reference implementations embed 48,000 to 69,000
residues per second, so a single pass over 2.6 billion proteins keeps an eight-GPU server busy for two to
three weeks.

We present HyPLM-Flash, a protein language model trained by Hyra 2.0, the research agent developed by our
team.
It is as accurate as the strongest encoders of its size, and both cheaper to train and faster to serve.
It retains the vocabulary and output head of ESM-2, making it a compelling and seamless alternative in existing pipelines. HyPLM-Flash is paired with a modern Transformer block and the Muon optimizer with a single pass over 2.6 billion curated and metagenomic sequences.
A dedicated inference engine then makes full use of the GPU through padding-free batching, fused kernels and INT8 matrix multiplication.

Trained on 0.73 times the data of ESM-2 650M and a fifth to an eighth of that of E1-600M and ESM~C 600M,
HyPLM-Flash matches these models on zero-shot fitness prediction, leads all self-supervised
single-sequence models on organismal fitness and gives the best long-range contact precision of all models
we evaluated.
Its engine embeds proteins 1.9 to 2.7 times faster than other encoders of the 600M class, which places
HyPLM-Flash on a new Pareto frontier of accuracy and speed (Figure~\ref{fig:teaser}).

Our contributions are:
\begin{itemize}
\item HyPLM-Flash, a 650M-parameter protein language model that reaches the accuracy of the strongest
encoders of its size with a fraction of their training data (Section~\ref{sec:model}).
\item An inference engine for H20 GPUs that embeds about 130,000 residues per second on one GPU
(Section~\ref{sec:engine}).
\item A matched evaluation of 28 public checkpoints from 13 model families on 32 tasks under a single
protocol with paired confidence intervals (Section~\ref{sec:evaluation}).
\end{itemize}

\section{Model and Training}\label{sec:model}

\subsection{Architecture}
\begin{figure}[t]
\centering
\begin{adjustbox}{max width=\linewidth}
\begin{tikzpicture}[node distance=4mm and 6mm, tall/.style={minimum height=12mm}]
  \node[hy/box, tall, font=\ttfamily\fontsize{8}{10}\selectfont] (tok) {M K T A Y \ldots};
  \node[hy/emb, tall, right=of tok] (emb) {Token embedding\\[-1pt]\hysub 33 $\times$ 1280};
  \node[hy/attn, tall, right=8mm of emb, minimum width=30mm] (blk) {\textbf{Transformer block}\\[-1pt]\hysub $\times$ 32 layers};
  \begin{scope}[on background layer]
    \foreach \s in {2.6pt,1.3pt}
      \draw[hy/attn, fill=hyblue!6, draw=hyblue!55] ($(blk.south west)+(\s,\s)$) rectangle ($(blk.north east)+(\s,\s)$);
  \end{scope}
  \node[hy/norm, tall, right=8mm of blk] (fn) {Final\\RMSNorm};
  \node[hy/box, tall, right=of fn] (head) {MLM head\\[-1pt]\hysub dense, GELU, LayerNorm\\\hysub linear $\rightarrow$ 33 logits};
  \foreach \a/\b in {tok/emb, emb/blk, blk/fn, fn/head} \draw[hy/arrow] (\a) -- (\b);

  \begin{scope}[yshift=-31mm]
    \node[hy/note, font=\fontsize{9.5}{10}\selectfont] (xin) {$x_\ell$};
    \node[hy/norm, right=7mm of xin] (n1) {RMSNorm};
    \node[hy/attn, tall, right=of n1] (att) {Self-attention\\[-1pt]\hysub 20 heads $\times$ 64\\\hysub QK-norm, RoPE};
    \node[hy/op, right=of att] (add1) {+};
    \node[hy/norm, right=7mm of add1] (n2) {RMSNorm};
    \node[hy/mlp, tall, right=of n2] (mlp) {SwiGLU MLP\\[-1pt]\hysub 1280 $\rightarrow$ 2 $\times$ 3584\\\hysub $\rightarrow$ 1280};
    \node[hy/op, right=of mlp] (add2) {+};
    \node[hy/note, right=6mm of add2, font=\fontsize{9.5}{10}\selectfont] (xout) {$x_{\ell+1}$};
  \end{scope}
  \foreach \a/\b in {xin/n1, n1/att, att/add1, add1/n2, n2/mlp, mlp/add2, add2/xout} \draw[hy/arrow] (\a) -- (\b);
  \coordinate (r1) at ($(xin.east)!0.5!(n1.west)$);
  \coordinate (r2) at ($(add1.east)!0.5!(n2.west)$);
  \draw[hy/arrow, rounded corners=3pt] (r1) -- ++(0,8.5mm) -| (add1.north);
  \draw[hy/arrow, rounded corners=3pt] (r2) -- ++(0,8.5mm) -| (add2.north);
  \fill[hyink!75] (r1) circle (0.9pt) (r2) circle (0.9pt);
  \begin{scope}[on background layer]
    \node[hy/group, fit=(xin)(xout)(att)(mlp), inner xsep=4mm, inner ysep=6mm, yshift=2.5mm] (grp) {};
  \end{scope}
  \draw[hyline, line width=0.6pt, dash pattern=on 2pt off 1.5pt]
    (blk.south west) -- (grp.north west) (blk.south east) -- (grp.north east);
\end{tikzpicture}
\end{adjustbox}
\caption{\textbf{Architecture of HyPLM-Flash.} Top: tokens from the 33-symbol ESM-2 alphabet are
embedded, processed by 32 identical Transformer blocks and decoded by a masked-language-modeling head.
Bottom: each block applies RMSNorm before bidirectional self-attention (per-head RMSNorm on queries and
keys, rotary position embeddings) and before a SwiGLU MLP, each inside a residual connection.}
\label{fig:architecture}
\end{figure}
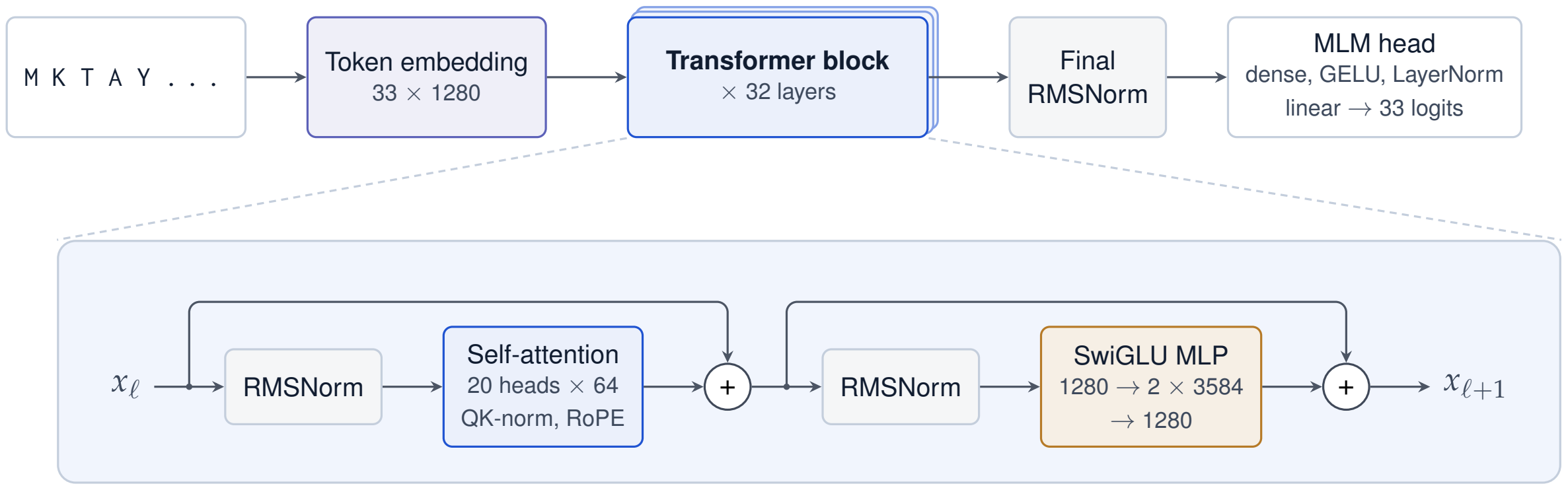

HyPLM-Flash is a bidirectional Transformer encoder~\citep{vaswani2017attention} trained with masked
language modeling~\citep{devlin2019bert}.
It uses the 33-token alphabet of ESM-2 and a RoBERTa-style output head~\citep{liu2019roberta,lin2023evolutionary},
and modernizes the Transformer block with pre-normalization by RMSNorm~\citep{zhang2019root}, rotary
position embeddings~\citep{su2021roformer}, per-head query--key
normalization~\citep{henry2020query,dehghani2023scaling} and SwiGLU feed-forward
layers~\citep{shazeer2020glu} (Figure~\ref{fig:architecture}).
Query--key normalization bounds the attention logits and keeps training stable at the high learning rates
used with Muon.
The model has 32 layers of width 1,280 with 20 attention heads and a SwiGLU width of 3,584, for a total of
652 million parameters, and reads proteins of up to 8,190 residues in a single window.

\subsection{Training}
\paragraph{Data.}
The training corpus combines UniRef90 with the UniRef50 representatives it does not
contain~\citep{suzek2015uniref,uniprot2025}, MGnify90~\citep{richardson2023mgnify}, the active entries of
UniParc and the non-redundant isolate proteomes of JGI IMG~\citep{chen2023imgm}.
We keep sequences of 30 to 8,192 residues, remove exact duplicates across sources and remove, by exact
match, the sequences of the benchmarks used during development, including all ProteinGym wild types and
variants.
The resulting 2.6 billion sequences (659 billion residues) are sampled in proportion to their length, so
that metagenomic and archival sequences supply about 80\% of the training tokens, and each source is seen
about once.

\paragraph{Objective and optimization.}
We use the standard masking recipe: 15\% of the residues are selected, of which 80\% are replaced by a mask
token, 10\% by a random amino acid and 10\% are left unchanged.
The weight matrices of the Transformer blocks are trained with Muon~\citep{jordan2024muon}, using decoupled
weight decay and an update scale matched to AdamW~\citep{liu2025muon}, and all other parameters with
AdamW~\citep{loshchilov2019decoupled}.
The learning rate follows a warmup--stable--decay schedule~\citep{hu2024minicpm,wen2024wsd}.

\paragraph{Training stages.}
Pretraining covers 713 billion tokens on crops of up to 1,022 residues.
A second stage of 13 billion tokens extends the crops to 8,190 residues and adds auxiliary structure
objectives on 1.6 million protein chains with experimental~\citep{berman2000pdb} or
predicted~\citep{lin2023evolutionary,varadi2022afdb} structures: a residue head predicts secondary
structure~\citep{kabsch1983dssp} and Foldseek 3Di states~\citep{vankempen2024foldseek}, and a pair head
predicts C$\beta$ distances~\citep{jumper2021alphafold}.
The auxiliary heads are used only during training.
Before training, we removed every structure chain with an MMseqs2 hit (E-value $\leq10^{-3}$, sequence
identity $\geq30\%$) to the test proteins of the structure and likelihood tasks or to a ProteinGym wild type.

\paragraph{Training compute.}
In total, HyPLM-Flash was trained on 726 billion tokens, or $2.8\times10^{21}$ FLOPs by the usual $6ND$
estimate~\citep{kaplan2020scaling,hoffmann2022training}.
This is 0.73 times the data of ESM-2 650M and 5.5 and 8.5 times less than E1-600M and ESM~C 600M, whose 4.0
and 6.2 trillion tokens amount to $1.5\times10^{22}$ and $2.1\times10^{22}$ FLOPs (Table~\ref{tab:compute}).

\begin{table}[t]
\caption{\textbf{Training data and compute of HyPLM-Flash and encoders of similar size.}}
\label{tab:compute}
\small
% Generated table; do not edit.
\begin{tabularx}{\textwidth}{@{}lNNNNL@{}}
\toprule
& & & \multicolumn{2}{c}{\colhead{Training FLOPs}} & \\
\cmidrule(lr){4-5}
\colhead{Model} & \colhead{Params} & \colhead{Tokens} & \colhead{6ND} & \colhead{Published} & \colhead{Corpus} \\
\midrule
\multicolumn{6}{@{}l}{\colhead{\textit{Encoders of similar size}}} \\
ESM-2 650M & 652M & 1.0T & \sci{3.9}{21} & \sci{1.6}{22}\rlap{\textsuperscript{a}} & UniRef50/90 (2021\_04) \\
ESM-1b & 652M & 66B & \sci{2.6}{20} & — & UniRef50 (2018\_03) \\
ESM C 600M & 575M & 6.2T & \sci{2.1}{22} & \sci{2.2}{22} & UniRef, MGnify, JGI (70\% clusters) \\
E1-600M & 641M & 4.0T & \sci{1.5}{22} & — & Profluent Protein Atlas, UniRef \\
AMPLIFY-350M & 354M & 2.5T\rlap{\textsuperscript{b}} & \sci{5.3}{21} & \sci{4.6}{21} & UniRef100, OAS, SCOP (UR100P) \\
DPLM-650M & 652M & 100B & \sci{4.3}{21}\rlap{\textsuperscript{c}} & — & UniRef50 \\
ESM3-open & 1.4B & 300B & \sci{2.5}{21} & — & UniRef, MGnify, JGI, structures \\
\midrule
\hyplmrow \textbf{HyPLM-Flash} & 652M & 726B & \sci{2.8}{21} & — & UniRef90/50, MGnify90, UniParc, JGI; 1.6M structures \\
\bottomrule
\end{tabularx}
\tablenote{6ND: total parameters $N$ times training tokens $D$; published FLOPs as reported by the authors; — not published. \textsuperscript{a}\,Estimated by the authors from GPU-days, not a 6ND count.\quad \textsuperscript{b}\,Upper bound: counts padded sequence slots.\quad \textsuperscript{c}\,DPLM-650M continues ESM-2 650M; tokens cover the final stage and FLOPs include the parent model (final stage alone: \sci{3.9}{20}).}
\tablenote{Sources: ESM-2 \citep{lin2023evolutionary,esm2discussion414}, ESM-1b \citep{rives2021biological,hf_esm1b_card}, ESM C \citep{candido2026language,esmc2024}, E1 \citep{jain2025e1}, AMPLIFY \citep{fournier2024protein}, DPLM \citep{wang2024diffusion}, ESM3 \citep{hayes2025simulating}.}

\end{table}

\section{Inference Engine}\label{sec:engine}
Embedding proteomes and scoring variant libraries are dominated by inference, and reference implementations
of protein language models leave much of a modern GPU idle.
We serve HyPLM-Flash with an inference engine written for NVIDIA H20 GPUs, which offer 148 TFLOPS in
BF16 and 296 TOPS in INT8 at 4.0 TB/s of memory bandwidth.

The engine rests on three ideas.
Proteins are packed into a single token stream without padding, which would otherwise fill 42\% of the
slots of a typical batch, and attention runs in the variable-length mode of
FlashAttention-3~\citep{shah2024flashattention3}.
Three Triton kernels~\citep{tillet2019triton} absorb every memory-bound operation: the residual addition
with RMSNorm, query--key normalization with the rotary embedding, and the SwiGLU activation, each fused with
the quantization of its output, while the residual stream stays in FP32.
Finally, the four projections of every block run as INT8 matrix multiplications.
Activations are scaled per token and per group of 128 channels, which confines the heavy-tailed outliers of
the feed-forward activations to their own group, and weights are scaled per output channel; each 128-wide
slice of the reduction accumulates in INT32 and is promoted to FP32 with its activation scale, the strategy
DeepSeek-V3 uses for FP8~\citep{deepseek2024v3}.

On a common workload of 2,048 held-out proteins, the engine embeds 131,000 residues per second on one H20.
This is 1.9 to 2.7 times faster than ESM~C 600M, ESM-1b, ESM-2 650M, DPLM-650M and E1-600M in their reference
implementations, and faster than ESM-2 150M, which has a quarter of the parameters (Figure~\ref{fig:teaser}a).
Accuracy in this report is evaluated with BF16 arithmetic.
INT8 keeps zero-shot fitness, structure, function annotation and likelihood within 0.01 of these results; for
supervised regression on near-identical protein variants, whose representations differ by only about 1\%,
we recommend the BF16 mode of the engine, which runs at 90,000 residues per second.

\section{Evaluation}\label{sec:evaluation}
We compare HyPLM-Flash with 28 public checkpoints from 13 model families, from ESM-2 8M to ESM-2 15B
(29 models in all), on 32 tasks.
All models run through the same evaluation code in BF16.
Zero-shot scores come from the masked-language-model head, and every supervised task fits a linear probe on
frozen representations from the layer, among six relative depths, that performs best on validation data.
Confidence intervals are paired bootstrap intervals over assays, proteins or sequences.
The main text focuses on encoders of similar size; Appendix~\ref{app:protocols} describes the protocols and
Appendix~\ref{app:results} reports every task for HyPLM-Flash and twelve other encoders.
E1~\citep{jain2025e1} is retrieval-augmented: it is trained with homologous sequences in its context and
improves when they are supplied, at the cost of an external sequence database, a homology search for every
query and longer inputs.
As for every other model, we run E1-600M on the query sequence alone.

\subsection{Zero-shot fitness prediction}\label{sec:proteingym}
We score each substitution by the masked-marginal log-likelihood ratio~\citep{meier2021esm1v} and report
the Spearman correlation over the 217 substitution assays of ProteinGym v1.3~\citep{notin2023proteingym}
with the benchmark's official aggregation.
Our runs of eight published ESM-2 and ESM~C checkpoints reproduce the leaderboard to within 0.0025.

\begin{figure}[t]
\centering\includegraphics[width=\linewidth]{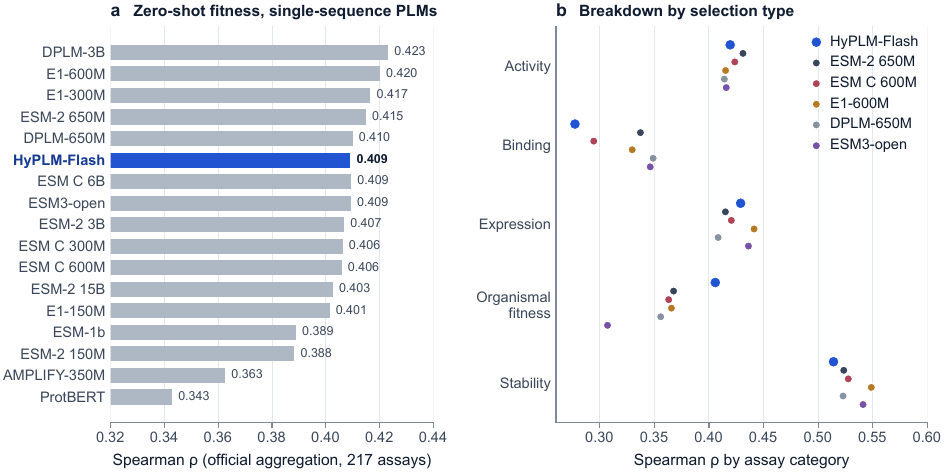}
\caption{\textbf{ProteinGym zero-shot fitness prediction.}
\textbf{a,} Average Spearman correlation over 217 substitution assays (official aggregation) for
single-sequence models; ESM3-open is scored from sequence alone.
\textbf{b,} Correlation by selection type for HyPLM-Flash and the main encoders of similar size.}
\label{fig:proteingym}
\end{figure}

HyPLM-Flash reaches a Spearman correlation of 0.409 (Figure~\ref{fig:proteingym}a).
Its differences to ESM-2 650M ($-0.006$, 95\% confidence interval $[-0.025, 0.010]$), ESM~C 600M ($+0.003$,
$[-0.015, 0.019]$), DPLM-650M ($-0.001$, $[-0.019, 0.015]$) and E1-600M ($-0.011$, $[-0.029, 0.005]$) are not
significant, and neither are those to ESM-2 3B, ESM-2 15B and ESM~C 6B, which were trained with 6 to 83 times
more compute.
Among the 55 single-sequence entries of the public leaderboard, HyPLM-Flash is tied for eighth place
(Table~\ref{tab:leaderboard}); the two entries at the top, VespaG and VESPA, are supervised predictors built
on PLM embeddings~\citep{marquet2022vespa,marquet2024vespag}.
By selection type (Figure~\ref{fig:proteingym}b), HyPLM-Flash scores 0.406 on organismal fitness, the
highest of all self-supervised single-sequence models on the leaderboard, and is on par with the other
encoders on activity and expression.

\subsection{Frozen-representation tasks}\label{sec:downstream}
Following TAPE~\citep{rao2019tape} and \citet{rao2020transformer}, we probe frozen representations with a
linear per-residue probe for secondary structure, a 32-dimensional bilinear head on residue pairs for
contacts, and linear probes on mean-pooled representations for all other tasks (Table~\ref{tab:downstream}).

\paragraph{Structure.}
HyPLM-Flash gives the best long-range contact precision of all 29 models: a precision at $L/5$ of
0.650 on ProteinNet, against 0.392 for ESM-2 650M, 0.434 for ESM~C 600M and 0.645 for E1-600M, and the
highest scores on the CASP14 and CASP15 targets as well (Appendix~\ref{app:results}), ahead of
ESM-2 15B and ESM~C 6B.
Its three-state secondary-structure accuracy on CB513 (0.882) is above that of ESM-2 650M (0.844) and
ESM~C 600M (0.851), on par with E1-600M (0.880), and exceeded only by ESM3-open (0.896) and ESM~C 6B (0.884)
among all models.

\paragraph{Function, localization and fitness.}
On enzyme and Gene Ontology annotation, HyPLM-Flash is within one point of the best encoder of its size
(EC $F_{\max}$ 0.885, GO-MF 0.661).
On fitness and property regression it ranks first among the models of Table~\ref{tab:downstream} on AAV
(two-vs-many, $\rho = 0.650$) and amylase (0.696), and second on Meltome (0.672) and fluorescence (0.669).
Averaged over the ten regression splits used for summary statistics (Appendix~\ref{app:protocols}), it scores
0.606, the highest of these models, ahead of E1-600M (0.597), ESM~C 600M (0.587), ESM-2 650M (0.572) and the
4.4 times larger ESM-2 3B (0.573).

\begin{table}[t]
\caption{\textbf{Frozen-representation tasks.}
Test metrics at the validation-selected layer: three-state accuracy on CB513 (SS), long-range contact
precision at $L/5$ on ProteinNet, fold accuracy, $F_{\max}$ for EC and GO molecular function, 10-class
localization accuracy, and Spearman $\rho$ for AAV (two-vs-many), amylase (one-to-many), fluorescence,
Meltome (mixed split) and stability. Bold marks the best value in each column.}
\label{tab:downstream}
\begin{tabular}{@{}lNNNNNNNNNNN@{}}
    \toprule
    & \multicolumn{3}{c}{\colhead{Structure}} & \multicolumn{3}{c}{\colhead{Function}} & \multicolumn{5}{c}{\colhead{Fitness and properties}} \\
    \cmidrule(lr){2-4}\cmidrule(lr){5-7}\cmidrule(lr){8-12}
    \colhead{Model} & \colhead{SS (Q3)} & \colhead{Contact} & \colhead{Fold} & \colhead{EC} & \colhead{GO-MF} & \colhead{Loc.} & \colhead{AAV} & \colhead{Amylase} & \colhead{Fluor.} & \colhead{Meltome} & \colhead{Stability} \\
    \midrule
    ESM-2 650M & 0.844 & 0.392 & 0.373 & 0.888 & \best{0.667} & 0.806 & 0.518 & 0.640 & 0.664 & 0.633 & 0.639 \\
    ESM C 600M & 0.851 & 0.434 & 0.345 & 0.886 & 0.665 & 0.808 & 0.551 & 0.681 & 0.663 & 0.670 & 0.706 \\
    E1-600M & 0.880 & 0.645 & 0.373 & 0.881 & 0.661 & \best{0.839} & 0.583 & 0.579 & 0.660 & \best{0.690} & 0.712 \\
    DPLM-650M & 0.842 & 0.394 & 0.370 & 0.882 & 0.663 & 0.806 & 0.569 & 0.674 & 0.668 & 0.623 & 0.746 \\
    ESM-1b & 0.814 & 0.410 & 0.362 & \best{0.889} & 0.659 & 0.793 & 0.544 & 0.653 & 0.659 & 0.593 & 0.596 \\
    AMPLIFY-350M & 0.797 & 0.210 & 0.286 & 0.865 & 0.631 & 0.729 & 0.574 & 0.592 & 0.660 & 0.585 & 0.579 \\
    ESM3-open & \best{0.896} & 0.568 & \best{0.408} & 0.880 & 0.664 & 0.750 & 0.551 & 0.618 & \best{0.669} & 0.573 & \best{0.758} \\
    \midrule
    \hyplmrow
    \textbf{HyPLM-Flash} & 0.882 & \best{0.650} & 0.322 & 0.885 & 0.661 & 0.794 & \best{0.650} & \best{0.696} & 0.669 & 0.672 & 0.687 \\
    \bottomrule
  \end{tabular}

\end{table}

\subsection{Masked-residue likelihood}\label{sec:likelihood}
We compute the negative log-likelihood (NLL) of masked residues with identical masks for all models on two
evaluation sets: 479 proteins from our held-out UniRef split, and 887 PDB chains released after 2 September 2026,
later than the training data of every model evaluated, clustered at 50\% identity with
MMseqs2~\citep{steinegger2017mmseqs2} after removing exact matches to our training corpus.
HyPLM-Flash has a lower NLL than ESM-2 650M, ESM~C 600M and E1-600M on both validation sets
(Table~\ref{tab:likelihood}).
On the recent PDB chains it is also ahead of ESM-2 3B and ESM-2 15B (1.148 against 1.327 and 1.162); of all
models for which we report a likelihood, only ESM~C 6B and ProtBERT score lower.

\begin{table}[t]
\caption{\textbf{Masked-residue likelihood.}
Negative log-likelihood (NLL, nats) and perplexity (PPL) on held-out UniRef proteins and on PDB chains released
after 2 September 2026. Lower is better.}
\label{tab:likelihood}
\centering\small
\begin{tabular}{@{}lNNNN@{}}
    \toprule
    & \multicolumn{2}{c}{\colhead{Held-out UniRef proteins}} & \multicolumn{2}{c}{\colhead{Recent PDB chains}} \\
    \cmidrule(lr){2-3}\cmidrule(l){4-5}
    \colhead{Model} & \colhead{NLL} & \colhead{PPL} & \colhead{NLL} & \colhead{PPL} \\
    \multicolumn{5}{@{}l}{\textit{Encoders of similar size}} \\
    ESM-2 650M & 1.820 & 6.17 & 1.591 & 4.91 \\
    ESM C 600M & 1.720 & 5.59 & 1.478 & 4.39 \\
    E1-600M & 1.811 & 6.11 & 1.560 & 4.76 \\
    AMPLIFY-350M & 1.743 & 5.72 & 1.155 & 3.17 \\
    DPLM-650M & 1.961 & 7.10 & 1.796 & 6.03 \\
    ProtBERT & 1.754 & 5.77 & 1.105 & 3.02 \\
    \midrule
    \multicolumn{5}{@{}l}{\textit{Larger encoders}} \\
    ESM-2 3B & 1.667 & 5.30 & 1.327 & 3.77 \\
    ESM-2 15B & 1.561 & 4.76 & 1.162 & 3.19 \\
    ESM C 6B & \best{1.414} & \best{4.11} & \best{1.054} & \best{2.87} \\
    \midrule
    \hyplmrow
    \textbf{HyPLM-Flash} & 1.698 & 5.46 & 1.148 & 3.15 \\
    \bottomrule
  \end{tabular}

\end{table}

\section{Conclusion}\label{sec:conclusion}
HyPLM-Flash shows that a protein language model can reach the accuracy of the strongest encoders of its
size with a fraction of their training data, and that an inference stack built around the target GPU makes
it the fastest of them.
It matches ESM-2 650M, ESM~C 600M and E1-600M on zero-shot fitness prediction, leads all self-supervised
single-sequence models on organismal fitness, gives the best contact precision in our
evaluation and models held-out and recent sequences better than ESM-2 650M, ESM~C 600M and E1-600M, while
embedding proteins 1.9 to 2.7 times faster than these models.
Together, these results place HyPLM-Flash on a new Pareto frontier of accuracy and efficiency.

\paragraph{Outlook.}
HyPLM-Flash is the first release of the HyPLM family.
Hyra 2.0 continues to develop it, and we will release updated versions as data and training recipes improve.
We are also training HyPLM-Pro, a larger model aimed at the highest accuracy rather than speed.
Like other single-sequence models, HyPLM-Flash reads sequence alone; conditioning on structure or on
retrieved homologs~\citep{su2024saprot,jain2025e1} is a natural direction for future versions.

\section*{Author Information}
\phantomsection\addcontentsline{toc}{section}{Author Information}
Haowei Lin, Cai Zhou and Xiangxin Zhou. All authors are affiliated with Tencent Hunyuan.

\bibliographystyle{plainnat}
\bibliography{refs}

\clearpage
\appendix
\section{Evaluation Protocols}\label{app:protocols}

\paragraph{Models.}
Baselines are public checkpoints run in BF16 through thin adapters around their reference implementations:
Hugging Face Transformers for ESM-1b, ESM-2, DPLM, SaProt, ProtBERT, ProtT5 and Ankh; the \code{esm} package
for ESM~C and ESM3; and the authors' code for E1 (single-sequence mode, without retrieved context sequences),
AMPLIFY and xTrimoPGLM.
Inputs longer than a model's window are processed in windows with 50\% overlap whose residue representations
are averaged.

\paragraph{Throughput.}
For the speed comparison (Figure~\ref{fig:teaser}a), all models embed the same 2,048 held-out proteins
(639,806 residues) in length-sorted batches of at most 16,384 tokens on one H20, using the fastest attention
kernel each implementation supports: FlashAttention-2 for ESM-1b, ESM-2 and DPLM, FlashAttention for ESM~C and
E1, xformers for AMPLIFY and PyTorch SDPA for ProtBERT and ESM3-open.
HyPLM-Flash runs on its INT8 engine.

\paragraph{Probes.}
Sequence-level tasks use representations mean-pooled over residues without special tokens; per-residue tasks
use the residue representations.
Features are cached in FP16, or in FP32 for the Ankh models, whose activations exceed the FP16 range.
For every task and model we evaluate six layers at relative depths 0.25, 0.5, 0.625, 0.75, 0.875 and 1.0 and
select the layer and the regularization strength on the validation split, or by 5-fold cross-validation on
the training split where a task has none (DGEB EC).
Regression probes are closed-form ridge regressions and classification probes are L2-regularized logistic
regressions fitted with L-BFGS; multi-label tasks fit one logistic output per term and report $F_{\max}$, and
the HumanPPI pair task uses the features $[a, b, a\odot b, |a-b|]$.
The contact head projects residue features to 32 dimensions and scores pairs with a symmetric bilinear form;
it is trained with AdamW on up to 8,000 ProteinNet proteins.
Contacts are residue pairs whose C$\beta$ atoms (C$\alpha$ for ProteinNet) are closer than 8~\AA, and long range
means a sequence separation of at least 24.
DGEB Arch retrieval is zero-shot, by cosine similarity between mean-pooled representations.

\paragraph{Zero-shot scores and likelihood.}
A variant's ProteinGym score is the sum over its substitutions of $\log p(\text{mutant}) - \log
p(\text{wild type})$, with each position masked on its own in the wild-type sequence and, for long targets, one
window centered on the position.
Per-assay Spearman correlations are rounded to three decimals and aggregated with the official procedure.
The validation sets use fixed masks (15\% of the canonical residues, one seeded draw per protein) and
identical crops for every model.
Models without a suitable masked-language-model head (ProtT5, Ankh, Ankh3) are omitted from both.

\paragraph{Statistics.}
Confidence intervals use 5,000 bootstrap replicates over assays (recomputing the official aggregation for each
replicate), proteins (structure and likelihood), queries (retrieval), or sequences or pairs; paired differences
use identical resamples.
Three regression splits, GB1 two-vs-rest and three-vs-rest and the cross-protein hydrolase split of FLIP2, are
reported in Tables~\ref{tab:full1} and~\ref{tab:full2} but left out of summary statistics (Figure~\ref{fig:teaser}
and the regression average of Section~\ref{sec:downstream}): their validation data come from the training
distribution while the test sets are strongly shifted, so probe selection on them is dominated by noise, and
numerically equivalent runs of the same model differ by up to 0.2.

\paragraph{Deviations.}
We could not reproduce the published ProteinGym scores of xTrimoPGLM: our runs fall 0.07 to 0.09 below the
leaderboard, and its likelihood is implausibly high, which points to an incompatibility between our adapter
and its masked-language-model head; we therefore withhold its zero-shot and likelihood results, while its
embedding-based results are unaffected.
ESM3-open and SaProt are evaluated from sequence alone, whereas their published ProteinGym results also use
structure inputs.

\begin{table}[t]
\caption{\textbf{Evaluation tasks.} Split sizes in proteins (ProteinGym: assays; CASP: domains).}
\label{tab:tasks}
\small
\begin{tabularx}{\textwidth}{@{}llrL@{}}
\toprule
Task & Metric & Train / valid / test & Source \\
\midrule
\multicolumn{4}{@{}l}{\textit{Zero-shot}} \\
ProteinGym substitutions & Spearman $\rho$ & 217 assays & ProteinGym v1.3~\citep{notin2023proteingym} \\
Validation set & NLL & -- / -- / 479 & held-out UniRef split \\
PDB time split & NLL & -- / -- / 887 & RCSB PDB entries released 9 Sep to 7 Oct 2026 \\
\midrule
\multicolumn{4}{@{}l}{\textit{Structure}} \\
Secondary structure (Q3) & accuracy & 8,678 / 2,170 / 513 & TAPE~\citep{rao2019tape}; CB513, TS115, CASP12 \\
Contacts, ProteinNet & P@$L/5$ long & 25,299 / 224 / 40 & TAPE, ProteinNet CASP12 \\
Contacts, CASP14 / CASP15 & P@$L/5$ long & -- / -- / 82, 95 & CASP evaluation domains \\
Remote homology & fold accuracy & 12,312 / 736 / 718 & TAPE (SCOP) \\
\midrule
\multicolumn{4}{@{}l}{\textit{Function, localization and interaction}} \\
EC (two splits) & $F_{\max}$ & 13,090 / 1,465 / 1,604 & SaProt EC split; DeepFRI EC split \\
GO-MF, GO-BP, GO-CC & $F_{\max}$ & 26,225 / 2,904 / 3,350 & DeepFRI GO split as packaged by SaProt \\
DGEB EC & macro F1 & 512 / -- / 128 & DGEB~\citep{westroberts2024dgeb} \\
DGEB Arch retrieval & MAP@5 & 2,343 queries & DGEB \\
DeepLoc (10-class, binary) & accuracy & 8,743 / 2,190 / 2,745 & DeepLoc 1.0 via PEER~\citep{xu2022peer} \\
Metal-ion binding & accuracy & 5,068 / 662 / 665 & as packaged by SaProt \\
HumanPPI & accuracy & 26,319 / 234 / 180 & PEER \\
\midrule
\multicolumn{4}{@{}l}{\textit{Fitness and property regression}} \\
AAV (two-vs-many, low-vs-high) & Spearman $\rho$ & 28,626 / 3,181 / 50,776 & FLIP~\citep{dallago2021flip} \\
GB1 (three splits) & Spearman $\rho$ & 381 / 43 / 8,309 & FLIP \\
Meltome (human cell, mixed) & Spearman $\rho$ & 5,149 / 643 / 1,366 & FLIP \\
Amylase (one-to-many, by mutation count) & Spearman $\rho$ & 2,574 / 644 / 488 & FLIP2 \\
Hydrolase to P06241 & Spearman $\rho$ & 11,970 / 2,993 / 9,972 & FLIP2 (cross-protein) \\
Fluorescence, stability & Spearman $\rho$ & 21,446 / 5,362 / 27,217 & TAPE \\
Thermostability & Spearman $\rho$ & 5,056 / 639 / 1,336 & Meltome human cell as packaged by SaProt \\
\bottomrule
\end{tabularx}
\tablenote{Where a row groups several tasks, sizes are for the first.}
\end{table}

\section{Additional Results}\label{app:results}
Table~\ref{tab:leaderboard} lists the top of the ProteinGym leaderboard, and Tables~\ref{tab:full1}
and~\ref{tab:full2} the primary metric of every task for HyPLM-Flash and twelve other encoders.

\begin{table}[t]
\caption{\textbf{Top of the ProteinGym zero-shot leaderboard} (substitutions, single-sequence models, Spearman
$\rho$ by selection type). Models are ranked by their average at the precision shown, and tied models share a
rank. $\dagger$ our evaluation; other rows are published leaderboard values (ProteinGym repository, commit
144fe22). VespaG and VESPA are supervised~\citep{marquet2022vespa,marquet2024vespag}.}
\label{tab:leaderboard}
\centering\small
\begin{tabular}{@{}rlNNNNNN@{}}
    \toprule
    \colhead{Rank} & \colhead{Model} & \colhead{Activity} & \colhead{Binding} & \colhead{Expression}
      & \colhead{\makecell{Organismal\\fitness}} & \colhead{Stability} & \colhead{Average} \\
    \midrule
    1 & VespaG & \best{0.493} & \best{0.370} & \best{0.456} & 0.437 & 0.533 & \best{0.458} \\
    2 & VESPA & 0.468 & 0.366 & 0.404 & \best{0.439} & 0.500 & 0.435 \\
    3 & DPLM-3B\textsuperscript{\dag} & 0.427 & 0.346 & 0.422 & 0.398 & 0.523 & 0.423 \\
    4 & E1-600M\textsuperscript{\dag} & 0.415 & 0.330 & 0.441 & 0.366 & \best{0.549} & 0.420 \\
    5 & E1-300M\textsuperscript{\dag} & 0.438 & 0.332 & 0.429 & 0.346 & 0.537 & 0.417 \\
    6 & ESM2 (650M) & 0.425 & 0.337 & 0.415 & 0.368 & 0.523 & 0.414 \\
    7 & DPLM-650M\textsuperscript{\dag} & 0.414 & 0.349 & 0.408 & 0.356 & 0.523 & 0.410 \\
    \hyplmrow
    8 & \textbf{HyPLM-Flash}\textsuperscript{\dag} & 0.419 & 0.277 & 0.429 & 0.406 & 0.514 & 0.409 \\
    8 & ESM C 6B\textsuperscript{\dag} & 0.428 & 0.309 & 0.417 & 0.374 & 0.518 & 0.409 \\
    10 & ESM-1v (ensemble) & 0.420 & 0.320 & 0.429 & 0.386 & 0.477 & 0.406 \\
    10 & ESM-C (300M) & 0.423 & 0.315 & 0.408 & 0.360 & 0.526 & 0.406 \\
    10 & ESM2 (3B) & 0.417 & 0.321 & 0.403 & 0.378 & 0.509 & 0.406 \\
    13 & ESM-C (600M) & 0.423 & 0.294 & 0.420 & 0.362 & 0.528 & 0.405 \\
    14 & E1-150M\textsuperscript{\dag} & 0.426 & 0.325 & 0.420 & 0.304 & 0.532 & 0.401 \\
    \bottomrule
  \end{tabular}

\end{table}

\begin{table}[t]
\caption{\textbf{All tasks, part 1:} primary test metric at the validation-selected layer for HyPLM-Flash and
the main encoders of similar size. Bold marks the best value in each row.}
\label{tab:full1}
\centering\footnotesize
\begin{tabular}{@{}llNNNNNNN@{}}
    \toprule
    \colhead{Task} & \colhead{Metric} & \colhead{ESM-2 650M} & \colhead{ESM-1b} & \colhead{ESM C 600M} & \colhead{E1-600M} & \colhead{AMPLIFY-350M} & \colhead{DPLM-650M} & \colhead{\textbf{HyPLM-Flash}} \\
    \midrule
    \multicolumn{9}{@{}l}{\textit{Zero-shot and likelihood}} \\
    ProteinGym zero-shot & Spearman $\rho$ & 0.415 & 0.389 & 0.406 & \best{0.420} & 0.363 & 0.410 & 0.409 \\
    NLL, held-out UniRef panel & NLL & 1.820 & 1.862 & 1.720 & 1.811 & 1.743 & 1.961 & \best{1.698} \\
    NLL, recent PDB chains & NLL & 1.591 & 1.502 & 1.478 & 1.560 & 1.155 & 1.796 & \best{1.148} \\
    \midrule
    \multicolumn{9}{@{}l}{\textit{Structure}} \\
    Secondary structure (CB513) & Q3 accuracy & 0.844 & 0.814 & 0.851 & 0.880 & 0.797 & 0.842 & \best{0.882} \\
    Contacts, ProteinNet & P@$L/5$, long range & 0.392 & 0.410 & 0.434 & 0.645 & 0.210 & 0.394 & \best{0.650} \\
    Contacts, CASP14 & P@$L/5$, long range & 0.366 & 0.322 & 0.385 & 0.537 & 0.198 & 0.367 & \best{0.556} \\
    Contacts, CASP15 & P@$L/5$, long range & 0.391 & 0.404 & 0.440 & 0.566 & 0.244 & 0.394 & \best{0.574} \\
    Remote homology (fold) & accuracy & \best{0.373} & 0.362 & 0.345 & \best{0.373} & 0.286 & 0.370 & 0.322 \\
    \midrule
    \multicolumn{9}{@{}l}{\textit{Function, localization and interaction}} \\
    EC, SaProt split & $F_{\max}$ & 0.888 & \best{0.889} & 0.886 & 0.881 & 0.865 & 0.882 & 0.885 \\
    EC, DeepFRI split & $F_{\max}$ & 0.874 & \best{0.878} & 0.876 & 0.871 & 0.861 & 0.874 & 0.874 \\
    GO-MF & $F_{\max}$ & \best{0.667} & 0.659 & 0.665 & 0.661 & 0.631 & 0.663 & 0.661 \\
    GO-BP & $F_{\max}$ & 0.496 & 0.485 & \best{0.499} & 0.497 & 0.467 & 0.496 & 0.492 \\
    GO-CC & $F_{\max}$ & 0.535 & 0.520 & 0.533 & 0.519 & 0.524 & 0.530 & \best{0.538} \\
    EC classification (DGEB) & macro F1 & 0.665 & 0.673 & 0.701 & \best{0.715} & 0.669 & 0.682 & 0.687 \\
    Arch retrieval (DGEB) & MAP@5 & 0.120 & \best{0.206} & 0.035 & 0.042 & 0.057 & 0.194 & 0.086 \\
    Localization, 10-class & accuracy & 0.806 & 0.793 & 0.808 & \best{0.839} & 0.729 & 0.806 & 0.794 \\
    Localization, binary & accuracy & 0.907 & 0.922 & 0.912 & 0.910 & 0.891 & \best{0.922} & 0.906 \\
    Metal-ion binding & accuracy & 0.725 & 0.690 & 0.678 & 0.710 & \best{0.737} & 0.728 & 0.695 \\
    HumanPPI & accuracy & 0.856 & 0.867 & 0.856 & \best{0.900} & 0.811 & 0.833 & 0.817 \\
    \midrule
    \multicolumn{9}{@{}l}{\textit{Fitness and property regression}} \\
    AAV two-vs-many & Spearman $\rho$ & 0.518 & 0.544 & 0.551 & 0.583 & 0.574 & 0.569 & \best{0.650} \\
    AAV low-vs-high & Spearman $\rho$ & 0.279 & 0.280 & 0.296 & 0.295 & 0.254 & 0.286 & \best{0.296} \\
    GB1 two-vs-rest & Spearman $\rho$ & 0.552 & 0.615 & 0.640 & \best{0.660} & 0.497 & 0.584 & 0.628 \\
    GB1 three-vs-rest & Spearman $\rho$ & 0.801 & 0.807 & 0.762 & 0.765 & 0.774 & 0.777 & \best{0.808} \\
    GB1 low-vs-high & Spearman $\rho$ & 0.377 & 0.417 & 0.395 & \best{0.474} & 0.273 & 0.404 & 0.469 \\
    Meltome human cell & Spearman $\rho$ & 0.696 & 0.698 & 0.697 & \best{0.720} & 0.665 & 0.684 & 0.684 \\
    Meltome mixed & Spearman $\rho$ & 0.633 & 0.593 & 0.670 & \best{0.690} & 0.585 & 0.623 & 0.672 \\
    Amylase one-to-many & Spearman $\rho$ & 0.640 & 0.653 & 0.681 & 0.579 & 0.592 & 0.674 & \best{0.696} \\
    Amylase by mutation count & Spearman $\rho$ & \best{0.575} & 0.503 & 0.512 & 0.539 & 0.298 & 0.502 & 0.553 \\
    Hydrolase to P06241 & Spearman $\rho$ & 0.172 & $-$0.224 & 0.203 & $-$0.295 & 0.212 & \best{0.380} & $-$0.192 \\
    Fluorescence & Spearman $\rho$ & 0.664 & 0.659 & 0.663 & 0.660 & 0.660 & 0.668 & \best{0.669} \\
    Stability & Spearman $\rho$ & 0.639 & 0.596 & 0.706 & 0.712 & 0.579 & \best{0.746} & 0.687 \\
    Thermostability & Spearman $\rho$ & 0.697 & 0.699 & 0.702 & \best{0.721} & 0.664 & 0.684 & 0.684 \\
    \bottomrule
  \end{tabular}

\end{table}

\begin{table}[t]
\caption{\textbf{All tasks, part 2:} other encoders. -- not reported (Appendix~\ref{app:protocols}).}
\label{tab:full2}
\centering\footnotesize
\begin{tabular}{@{}llNNNNNN@{}}
    \toprule
    \colhead{Task} & \colhead{Metric} & \colhead{ESM3-open} & \colhead{ProtBERT} & \colhead{ProtT5-XL} & \colhead{Ankh-base} & \colhead{Ankh3-large} & \colhead{xTrimoPGLM-1B} \\
    \midrule
    \multicolumn{8}{@{}l}{\textit{Zero-shot and likelihood}} \\
    ProteinGym zero-shot & Spearman $\rho$ & \best{0.409} & 0.343 & -- & -- & -- & -- \\
    NLL, held-out UniRef panel & NLL & 1.920 & \best{1.754} & -- & -- & -- & -- \\
    NLL, recent PDB chains & NLL & 1.828 & \best{1.105} & -- & -- & -- & -- \\
    \midrule
    \multicolumn{8}{@{}l}{\textit{Structure}} \\
    Secondary structure (CB513) & Q3 accuracy & \best{0.896} & 0.785 & 0.844 & 0.847 & 0.810 & 0.799 \\
    Contacts, ProteinNet & P@$L/5$, long range & \best{0.568} & 0.229 & 0.566 & 0.546 & 0.383 & 0.321 \\
    Contacts, CASP14 & P@$L/5$, long range & 0.479 & 0.200 & \best{0.499} & 0.467 & 0.307 & 0.302 \\
    Contacts, CASP15 & P@$L/5$, long range & 0.521 & 0.250 & 0.521 & \best{0.523} & 0.369 & 0.352 \\
    Remote homology (fold) & accuracy & \best{0.408} & 0.270 & 0.326 & 0.362 & 0.288 & 0.364 \\
    \midrule
    \multicolumn{8}{@{}l}{\textit{Function, localization and interaction}} \\
    EC, SaProt split & $F_{\max}$ & 0.880 & 0.861 & \best{0.886} & 0.881 & 0.868 & 0.877 \\
    EC, DeepFRI split & $F_{\max}$ & 0.878 & 0.860 & \best{0.880} & 0.870 & 0.854 & 0.872 \\
    GO-MF & $F_{\max}$ & \best{0.664} & 0.622 & 0.660 & 0.659 & 0.640 & 0.649 \\
    GO-BP & $F_{\max}$ & 0.493 & 0.454 & \best{0.495} & 0.484 & 0.476 & 0.484 \\
    GO-CC & $F_{\max}$ & 0.508 & 0.529 & \best{0.533} & 0.502 & 0.512 & 0.524 \\
    EC classification (DGEB) & macro F1 & \best{0.681} & 0.589 & 0.671 & 0.680 & 0.645 & 0.502 \\
    Arch retrieval (DGEB) & MAP@5 & 0.277 & 0.072 & 0.193 & \best{0.311} & 0.270 & 0.124 \\
    Localization, 10-class & accuracy & 0.750 & 0.739 & \best{0.831} & 0.792 & 0.764 & 0.739 \\
    Localization, binary & accuracy & 0.900 & 0.894 & \best{0.919} & 0.906 & 0.918 & 0.906 \\
    Metal-ion binding & accuracy & 0.659 & 0.707 & 0.741 & \best{0.746} & 0.692 & 0.719 \\
    HumanPPI & accuracy & 0.822 & 0.833 & \best{0.889} & 0.833 & 0.839 & 0.839 \\
    \midrule
    \multicolumn{8}{@{}l}{\textit{Fitness and property regression}} \\
    AAV two-vs-many & Spearman $\rho$ & \best{0.551} & 0.409 & 0.533 & 0.539 & 0.496 & 0.322 \\
    AAV low-vs-high & Spearman $\rho$ & 0.287 & 0.231 & 0.253 & \best{0.296} & 0.267 & 0.270 \\
    GB1 two-vs-rest & Spearman $\rho$ & 0.519 & 0.507 & 0.414 & \best{0.617} & 0.299 & 0.592 \\
    GB1 three-vs-rest & Spearman $\rho$ & 0.766 & 0.765 & 0.781 & 0.796 & \best{0.814} & 0.770 \\
    GB1 low-vs-high & Spearman $\rho$ & 0.455 & 0.326 & 0.402 & 0.386 & \best{0.471} & 0.461 \\
    Meltome human cell & Spearman $\rho$ & 0.682 & 0.660 & 0.682 & 0.669 & 0.668 & \best{0.698} \\
    Meltome mixed & Spearman $\rho$ & 0.573 & 0.621 & \best{0.661} & 0.527 & 0.545 & 0.578 \\
    Amylase one-to-many & Spearman $\rho$ & 0.618 & 0.544 & \best{0.682} & 0.628 & 0.653 & 0.656 \\
    Amylase by mutation count & Spearman $\rho$ & 0.524 & 0.484 & 0.398 & \best{0.548} & 0.353 & 0.359 \\
    Hydrolase to P06241 & Spearman $\rho$ & \best{0.527} & 0.246 & 0.268 & 0.096 & $-$0.128 & 0.230 \\
    Fluorescence & Spearman $\rho$ & 0.669 & 0.641 & 0.663 & 0.642 & 0.669 & \best{0.673} \\
    Stability & Spearman $\rho$ & 0.758 & 0.516 & 0.586 & 0.649 & \best{0.759} & 0.660 \\
    Thermostability & Spearman $\rho$ & 0.679 & 0.661 & 0.680 & 0.670 & 0.684 & \best{0.696} \\
    \bottomrule
  \end{tabular}

\end{table}

\end{document}